\documentclass[conference]{IEEEtran}
\usepackage{float}

\usepackage{graphicx}
\usepackage{amsmath,amssymb,amsfonts}
\usepackage{algorithmic}
\usepackage{graphicx}
\usepackage{textcomp}
\usepackage[backend=biber, style=numeric, sorting=ynt]{biblatex}
\usepackage{xcolor}
\def\BibTeX{{\rm B\kern-.05em{\sc i\kern-.025em b}\kern-.08em
    T\kern-.1667em\lower.7ex\hbox{E}\kern-.125emX}}

\usepackage{pgfplots}
\usepackage{tikz}

\pgfplotsset{compat=1.18}

\begin{document}

\title{Neurosymbolics for Data Engineering: Achieving Over 50\% Token Reduction On Long Context Tasks Without Finetuning\\
}

\author{
\IEEEauthorblockN{Vishvesh Bhat}
\IEEEauthorblockA{
\textit{CoreThink AI} \\
San Jose, United States \\
vish@corethink.ai \\
vbhat@ucsd.edu
}

}

\maketitle

\begin{abstract}
Large Language Models (LLMs) are increasingly deployed for sophisticated data engineering tasks, such as generating structured queries from natural language (Text-to-SQL) and automating complex spreadsheet operations. However, maximizing their utility demands both higher, fine-tuning-free accuracy and solutions to the computational bottleneck imposed by the Transformer architecture's inherent quadratic $O(n^2)$ time complexity. This paper introduces a \textbf{novel, drop-in neurosymbolic layer} designed to seamlessly integrate into existing LLM backbones, enhancing logical reasoning and mitigating long-context resource consumption. On the reasoning front, the layer immediately and significantly improves performance, yielding an average accuracy increase of 8.5\% across rigorous benchmarks including \textbf{BIRD-CRITIC and LiveSQLBench}, critically achieving these gains \textbf{without any task-specific finetuning or RLHF}. 

Concurrently, we repurpose this approach to address the severe computational strain of long-context inference. By leveraging symbolic processing to prioritize and compress relevant contextual information, the layer reduces the effective token usage by \textbf{over 50\%} and brings the effective time complexity down from $O(n^2)$ to approximately $O(n)$ on certain long context tasks. This dual-impact approach not only makes LLMs substantially more reliable for data engineering but also drastically reduces the computational pressure on inference chips, making large-context tasks more manageable and cost-effective.
\end{abstract}


\section{Introduction}
The integration of Large Language Models (LLMs) into the data ecosystem has fundamentally transformed how users interact with and manipulate structured information. LLMs have demonstrated remarkable potential in bridging the gap between natural language and structured operations, most notably in Text-to-SQL generation and the automation of intricate spreadsheet functions \cite{mohammadjafari2025naturallanguagesqlreview}. These capabilities promise significant advancements in enterprise data accessibility and business intelligence, yet their wide-scale deployment is hampered by two critical limitations.

First, while LLMs excel at language fluency, they frequently exhibit a deficit in the logical and systematic reasoning required for complex data engineering tasks \cite{heesch2025evaluatinglargelanguagemodels}. Current state-of-the-art models often fail on highly compositional queries or subtle schema relationships, resulting in unreliable outputs on specialized and robust benchmarks like the \textbf{BIRD-Bench suite} and \textbf{SpreadsheetBench}. Achieving reliable performance typically necessitates expensive and iterative finetuning and reinforcement learning, which is not always practical. Second, the computational overhead of the ubiquitous Transformer architecture presents a severe bottleneck \cite{hilsenbek2024breakingattentionbottleneck}. The self-attention mechanism's quadratic time complexity, $O(n^2)$, is computationally taxing for high-capacity inference chips, especially as models are pushed to handle increasingly long contextual inputs. This inefficiency drives up operational costs and limits the scalability of LLM-based data pipelines.

To address this dual challenge of accuracy and efficiency, we propose a novel, drop-in neurosymbolic augmentation layer. This layer is designed to be model-agnostic, integrating seamlessly into any pre-trained LLM without requiring architectural modifications or expensive re-training. By explicitly injecting symbolic reasoning capabilities into the model's forward pass, we can enhance logical grounding and context utilization. \cite{liu2024neuralsymbolic}

Our results demonstrate a significant leap forward on both fronts. On logical reasoning benchmarks, the neurosymbolic layer acts as an immediate performance booster, yielding an average accuracy improvement of 8.5\% across \textbf{BIRD-CRITIC and LiveSQLBench}, critically achieving this without any fine-tuning (RLHF or otherwise) \cite{ma2024spreadsheetbench} \cite{zhu2025sheetmind} \cite{payan2023instructexcel}.  Simultaneously, by employing a symbolically-guided context compression mechanism, the layer dramatically optimizes runtime efficiency. We show that our approach reduces the effective token usage by over 50\% and effectively brings the model's practical time complexity down from $O(n^2)$ to approximately $O(n)$ on \textbf{Longbench v2}, thereby alleviating pressure on inference hardware\cite{shen2018efficient}.

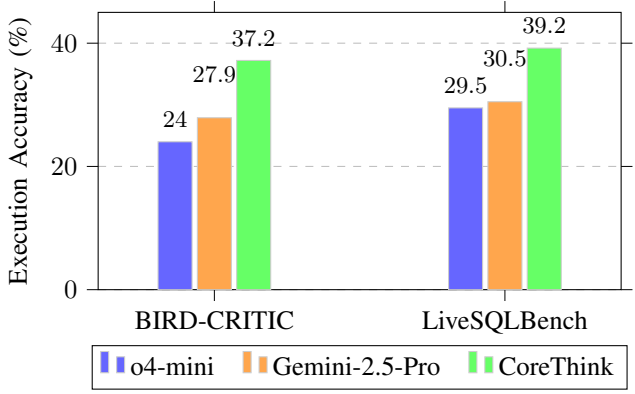
\begin{figure}[h]
\centering
\begin{tikzpicture}[xshift=-0.5cm] 
\begin{axis}[
    ybar,
    bar width=0.45cm,
    width=8.5cm,
    height=5.25cm,
    ylabel={Execution Accuracy (\%)},
    symbolic x coords={BIRD-CRITIC, LiveSQLBench},
    xtick=data,
    ymin=0,
    ymax=45,
    legend style={
        at={(0.5,-0.2)},
        anchor=north,
        legend columns=3,
        /tikz/every even column/.append style={column sep=0.3cm}
    },
    ymajorgrids=true,
    grid style=dashed,
    enlarge x limits=0.4,
    tick align=outside,
    tick style={black},
]

\addplot[fill=blue!60, draw=black!20,
    nodes near coords,
    every node near coord/.append style={font=\small, yshift=2pt}
] coordinates {(BIRD-CRITIC,24.0) (LiveSQLBench,29.5)};

\addplot[fill=orange!70, draw=black!20,
    nodes near coords,
    every node near coord/.append style={font=\small, yshift=10pt}
] coordinates {(BIRD-CRITIC,27.9) (LiveSQLBench,30.5)};

\addplot[fill=green!60, draw=black!20,
    nodes near coords,
    every node near coord/.append style={font=\small, yshift=2pt}
] coordinates {(BIRD-CRITIC,37.2) (LiveSQLBench,39.2)};

\legend{o4-mini, Gemini-2.5-Pro, CoreThink}

\end{axis}
\end{tikzpicture}
\caption{Execution accuracy on text-to-SQL benchmarks. CoreThink outperforms existing approaches on both BIRD-CRITIC and LiveSQLBench.}
\label{fig:text2sql}
\end{figure}

\section{Relevant Work}

\subsection{Text-to-SQL Benchmarks}

\textbf{BIRD-CRITIC} and \textbf{LiveSQLBench} represent extensions and evolutions of text-to-SQL evaluation \cite{li2025swesqlilluminatingllmpathways}. BIRD-CRITIC introduces a framework for iterative SQL refinement, where models can critique and correct their initial SQL generations based on execution feedback, simulating how developers actually debug queries. This addresses a key limitation of single-shot generation benchmarks by allowing models to learn from errors. LiveSQLBench, meanwhile, focuses on temporal robustness by testing whether text-to-SQL systems can handle queries about constantly updating real-world databases. It evaluates whether models can generate correct SQL for time-sensitive questions where database contents change frequently, such as sports statistics or stock prices. Together, these benchmarks push beyond static accuracy metrics to assess more dynamic, practical aspects of text-to-SQL systems including error correction, real-time data handling, and robust performance across diverse and evolving database environments.

\subsection{Long Context Benchmarks}

\textbf{LongBench v2}  is an advanced benchmark designed to evaluate large language models' capabilities in handling long-context inputs, extending and improving upon its predecessor with more challenging and diverse tasks \cite{bai2025longbenchv2deeperunderstanding}. Released as the field moved toward models supporting context windows of 100K tokens or more, LongBench v2 addresses limitations in earlier long-context benchmarks by incorporating harder tasks that require genuine long-range reasoning rather than simple retrieval. It includes multiple task categories such as single-document QA, multi-document QA, summarization, few-shot learning, code completion, and synthetic tasks that test specific abilities like "needle in a haystack" retrieval across extremely long contexts. The benchmark emphasizes tasks where the answer requires synthesizing information scattered throughout the entire context window, preventing models from succeeding through shortcut strategies or only attending to local information. LongBench v2 also incorporates more realistic document lengths and structures, including technical papers, books, and concatenated conversation histories that mirror real-world use cases.

\textbf{BFCL v3 Long Context} (Berkeley Function-Calling Leaderboard version 3 with Long Context support) extends function-calling evaluation to scenarios involving extensive context windows \cite{bfcl_v3}. Function calling—where models must decide when and how to invoke external tools or APIs based on user requests—becomes significantly more challenging when the relevant information for making these decisions is distributed across long documents or conversation histories. BFCL v3 Long Context tests whether models can accurately identify which functions to call, extract the correct parameters from thousands of tokens of context, and maintain coherent function-calling behavior across extended interactions \cite{kate2025longfunceval}. This benchmark is crucial for evaluating production-ready AI assistants that need to manage long conversations, reference extensive documentation, or work with large codebases while still making precise function calls. It addresses a critical gap between isolated function-calling benchmarks and the messy, context-rich environments where AI agents actually operate, ensuring models can maintain both long-range comprehension and precise tool use simultaneously.

\section{Our Approach: The General Symbolics Reasoning (GSR) Framework}
The \textbf{General Symbolics Reasoning (GSR) framework} represents a paradigm designed to perform stable, domain-adaptable, and computationally efficient reasoning entirely within natural language \cite{corethinkreasoner}. By operating on a pure natural language-to-natural language (NL-to-NL) basis, GSR avoids the representational loss and brittleness associated with translating human language into formal logic or high-dimensional vectors. The architecture employs a layered approach to systematically handle reasoning from input to explanation, preserving the full context and nuance of the original language. Each layer within this idealized architecture maintains a distinct responsibility in the reasoning pipeline.

\subsection{Native Language Parsing and Semantic Preservation}

The framework begins with direct natural language input, where all reasoning processes remain within natural language throughout, eliminating the need for intermediary formalisms. This design choice ensures that no semantic information is lost at the outset of the reasoning process. The system incorporates sophisticated ambiguity identification capabilities, utilizing word sense disambiguation and linguistic pattern recognition inherent to the language itself to resolve potential ambiguities.

\subsection{In-Language Reasoning Architecture}

GSR applies logical rules through natural language transformations that manipulate language components based on their syntactic and semantic relationships, rather than relying on abstract symbols. This approach to constraint enforcement through NL patterns preserves crucial contextual information. Unlike formal abstractions, GSR maintains pragmatic distinctions—such as the difference between ``must'' and ``should''—along with modality and specificity directly within the language, enabling more nuanced inference capabilities.

\subsection{Execution and Explainability}

Each step of the reasoning process remains human-interpretable through verbatim reasoning traces, exposing the exact reasoning path, intermediate conclusions, and any detected contradictions in plain, reviewable language. The framework surfaces inconsistencies through direct language annotations, such as highlighting conflicts in assumptions, rendering the entire process transparent and debuggable. This commitment to explainability ensures that error propagation can be traced and understood at every stage.

\subsection{Avoiding Representational Translation Pitfalls}

A fundamental principle of GSR is avoiding the loss inherent in translating natural language into vectors or formal logic. Such translations strip away crucial context, whereas an NL-to-NL process preserves all original information, leading to higher-fidelity reasoning. Natural language possesses far greater comprehensiveness and expressiveness than formal logic, which is inherently reductionist. Forcing natural language into predefined, rigid structures inevitably discards the rich contextual fabric of human expression. GSR leverages the expressiveness of language as its core strength, ensuring comprehensive reasoning capabilities.

\subsection{Computational Optimization Layer}

Despite its emphasis on preserving linguistic richness, GSR incorporates computational optimization mechanisms. Entity tagging and search-based pruning minimize extraneous inferences that might introduce noise to the reasoning trace. The framework is architected for real-time performance, designed to support long-horizon reasoning with high stability and without dependence on massive computational resources such as GPUs, making it practically deployable across diverse environments.

\subsection{A Neurosymbolic Step Towards GSR}

While the fully realized GSR framework described above remains the ultimate goal, this paper talks about a concrete and powerful step towards its implementation: a neurosymbolic framework whose design is directly inspired by GSR principles. This hybrid system serves as a practical bridge, approximating the pure NL-to-NL ideal by using a symbolic scaffold to orchestrate and compose smaller, efficient Large Language Models (LLMs).

In this architecture, the symbolic framework provides the structured, compositional reasoning path, while the neural components handle the nuanced, pattern-based tasks of parsing and transformation at each step. This approach allows us to achieve the core benefits of GSR—such as compositional logic and interpretable reasoning traces—in a practical system today. All results and analysis presented in this paper are based on this neurosymbolic implementation, which validates the core principles of GSR and marks a critical milestone toward achieving a truly comprehensive, in-language reasoning intelligence.

\section{Evaluation}

We evaluate the GSR neurosymbolic framework across two distinct domains to demonstrate its versatility, computational efficiency, and reasoning capabilities. Our experiments span structured query generation and long-context processing tasks, establishing new state-of-the-art results on multiple benchmarks while maintaining interpretability and efficiency.

\subsection{GSR for Text-to-SQL Generation}

\begin{figure}[!t]
    \centering
    \includegraphics[width=\columnwidth, height=10cm, keepaspectratio]{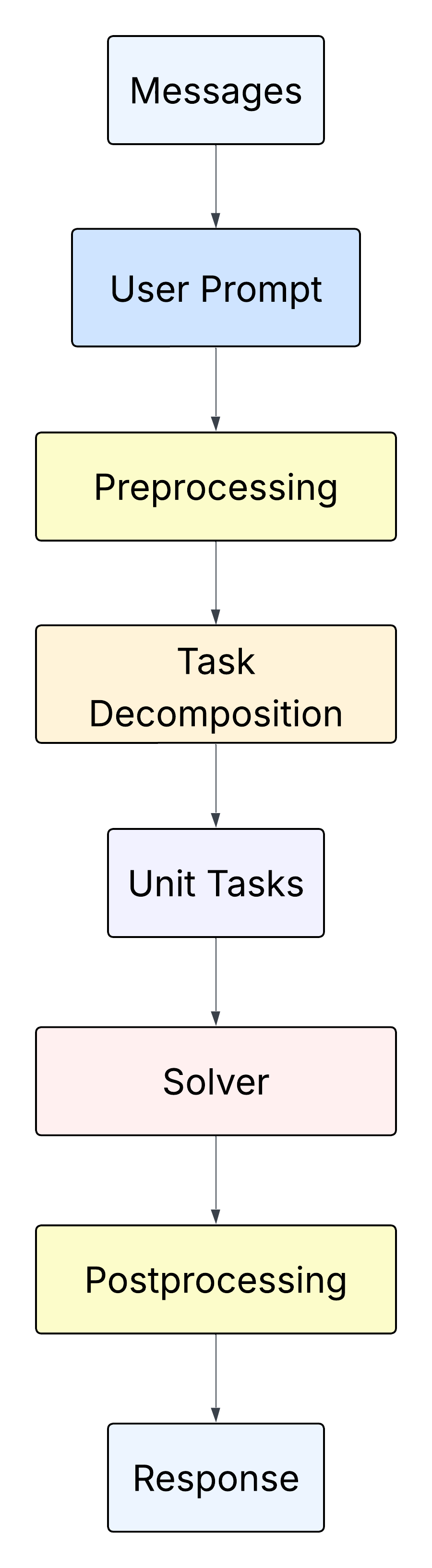}
    \caption{Overview of the GSR pipeline for Text-to-SQL Generation}
    \label{fig:evaluation-overview}
\end{figure}

We implemented GSR as a drop-in reasoning layer for text-to-SQL tasks, evaluating its performance on two challenging benchmarks: BIRD-CRITIC and LiveSQLBench. BIRD-CRITIC tests iterative SQL refinement capabilities with complex, real-world databases containing inconsistent data, while LiveSQLBench evaluates temporal robustness on constantly updating databases.

\subsubsection{Results}

Table~\ref{tab:text2sql} presents our results compared to existing state-of-the-art systems. GSR achieves top performance on both benchmarks, demonstrating superior reasoning and error correction capabilities.

\begin{table}[h]
\centering
\begin{tabular}{lcc}
\hline
\textbf{Model} & \textbf{BIRD-CRITIC} & \textbf{LiveSQLBench} \\
\hline
o4-mini & 24.0\% & 29.5\% \\
Gemini-2.5-Pro & 27.9\% & 30.5\% \\
\textbf{CoreThink} & \textbf{37.2\%} & \textbf{39.2\%} \\
\hline
\end{tabular}
\caption{Execution accuracy on text-to-SQL benchmarks. GSR outperforms existing approaches on both BIRD-CRITIC and LiveSQLBench.}
\label{tab:text2sql}
\end{table}

\subsubsection{Why This Matters}

The superior performance on these benchmarks demonstrates GSR's ability to handle complex, multi-step reasoning required for translating natural language into executable SQL queries. The iterative refinement capability tested by BIRD-CRITIC is particularly important for production systems, where initial query generation may require debugging based on execution feedback \cite{li2025swesqlilluminatingllmpathways}. LiveSQLBench results validate GSR's robustness to temporal dynamics and its ability to reason about time-sensitive data contexts \cite{li2025swesqlilluminatingllmpathways}. Unlike black-box neural approaches, GSR's reasoning traces provide full transparency into how queries are constructed, enabling developers to understand, validate, and debug the query generation process \cite{corethinkreasoner}. This interpretability is crucial for enterprise deployments where query correctness and auditability are paramount.

\subsection{GSR for Long-Context Processing}

\begin{figure}[!t]
    \centering
    \includegraphics[width=\columnwidth, height=15cm, keepaspectratio]{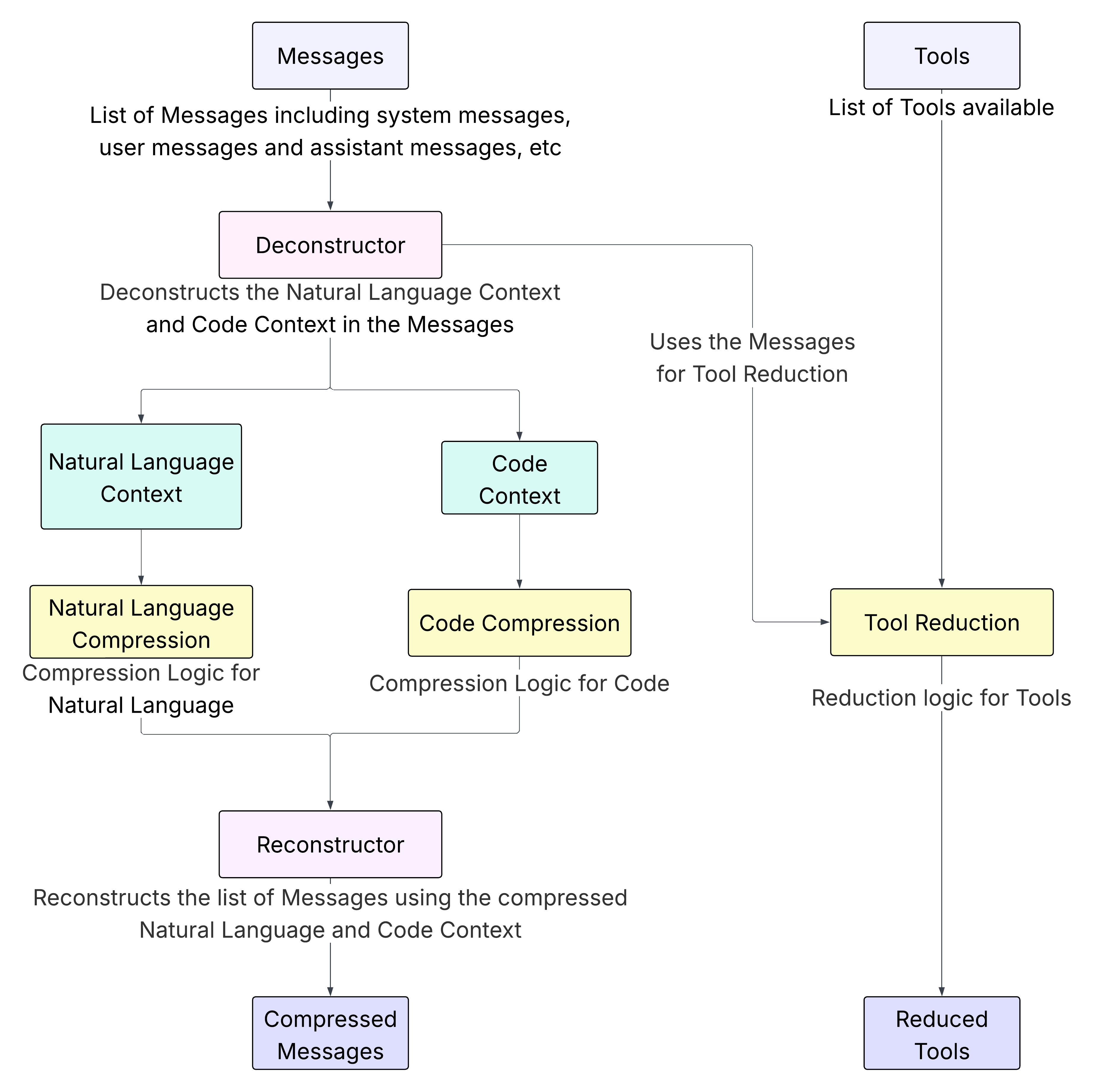}
    \caption{Overview of the GSR pipeline for Token Reduction}
    \label{fig:evaluation-overview-2}
\end{figure}

A key advantage of the GSR framework is its computational efficiency on long-context tasks \cite{corethinkreasoner}. The attention mechanism in transformer-based models scales as $O(n^2)$ with sequence length, creating substantial computational bottlenecks for long documents. GSR addresses this through intelligent input compression and structured reasoning, effectively reducing complexity closer to $O(n)$ on Longbench v2 by minimizing redundant token processing while maintaining reasoning quality.

We evaluate GSR on two long-context benchmarks: LongBench v2, which tests genuine long-range reasoning across multiple task categories, and BFCL v3 Long Context, which evaluates function-calling capabilities with extensive context windows.

\subsubsection{Results}

\begin{figure}[h]
\centering
\begin{tikzpicture}[xshift=-0.4cm]
\begin{axis}[
    ybar,
    bar width=0.45cm,
    width=8.5cm,
    height=6cm,
    ylabel={Percentage (\%)},
    symbolic x coords={Accuracy, Token Reduction},
    xtick=data,
    ymin=0,
    ymax=110, 
    legend style={
        at={(0.5,-0.2)},
        anchor=north,
        legend columns=2,
        /tikz/every even column/.append style={column sep=0.3cm}
    },
    ymajorgrids=true,
    grid style=dashed,
    enlarge x limits=0.4,
    tick align=outside,
    tick style={black},
    nodes near coords,
    nodes near coords style={font=\small, yshift=2pt},
]

\addplot[fill=blue!60, draw=black!20] coordinates {(Accuracy,65.4) (Token Reduction,0)};
\addplot[fill=orange!70, draw=black!20,
    every node near coord/.append style={yshift=8pt}
] coordinates {(Accuracy,63.2) (Token Reduction,92)};

\legend{Base (Deepseek R1), Deepseek R1 + NS}

\end{axis}
\end{tikzpicture}
\caption{Performance on LongBench v2. GSR maintains competitive accuracy (63.2\% vs 65.4\%) while achieving 92\% token reduction.}
\label{fig:longbench}
\end{figure}
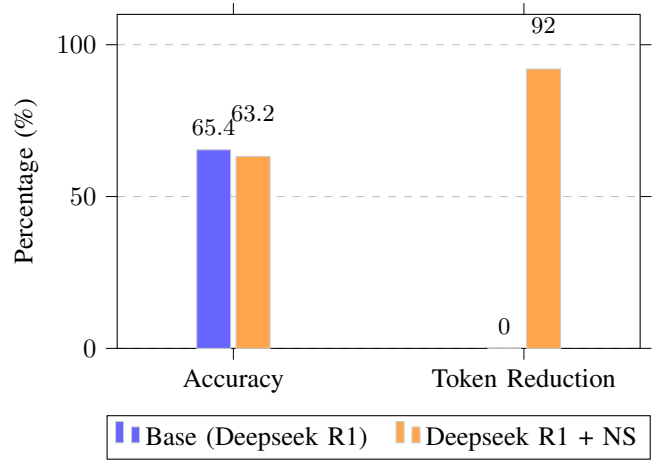

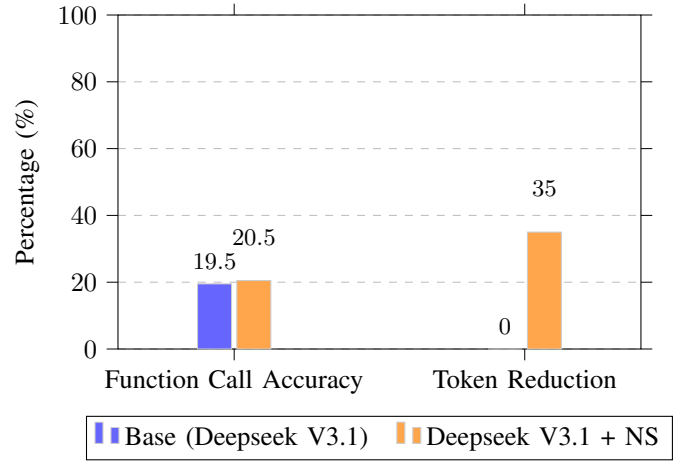
\begin{figure}[h]
\centering
\begin{tikzpicture}[xshift=-0.4cm] 
\begin{axis}[
    ybar,
    bar width=0.45cm,
    width=8.5cm,
    height=6cm,
    ylabel={Percentage (\%)},
    symbolic x coords={Function Call Accuracy, Token Reduction},
    xtick=data,
    ymin=0,
    ymax=100,
    legend style={
        at={(0.5,-0.2)},
        anchor=north,
        legend columns=2,
        /tikz/every even column/.append style={column sep=0.3cm}
    },
    ymajorgrids=true,
    grid style=dashed,
    enlarge x limits=0.4,
    tick align=outside,
    tick style={black},
    nodes near coords,
    nodes near coords style={font=\small, yshift=2pt},
]

\addplot[fill=blue!60, draw=black!20] coordinates {(Function Call Accuracy,19.5) (Token Reduction,0)};
\addplot[fill=orange!70, draw=black!20,
    every node near coord/.append style={yshift=8pt}
] coordinates {(Function Call Accuracy,20.5) (Token Reduction,35)};

\legend{Base (Deepseek V3.1), Deepseek V3.1 + NS}

\end{axis}
\end{tikzpicture}
\caption{Performance on BFCL v3 Long Context. GSR improves function-calling accuracy (20.5\% vs 19.5\%) with 35\% token reduction.}
\label{fig:bfcl}
\end{figure}

Tables~\ref{tab:longbench} and~\ref{tab:bfcl} present our results on long-context benchmarks, along with computational efficiency metrics.

\begin{table}[h]
\centering
\begin{tabular}{lcc}
\hline
\textbf{Model} & \textbf{Accuracy} & \textbf{Token reduction} \\
\hline
Base accuracy (Deepseek R1) & 65.4\% & - \\
\textbf{Deepseek R1 + Neurosymbolic} & \textbf{63.2\%} & \textbf{92\%} \\
\textbf{token reduction} & & \\
\hline
\end{tabular}
\caption{Performance on LongBench v2. GSR maintains competitive accuracy while significantly reducing token consumption and computational cost.}
\label{tab:longbench}
\end{table}

\begin{table}[h]
\centering
\begin{tabular}{lcc}
\hline
\textbf{Model} & \textbf{Function Call} & \textbf{Token reduction} \\
\hline
Base accuracy (Deepseek V3.1) & 19.5\% & - \\
\textbf{Deepseek V3.1 + Neurosymbolic} & \textbf{20.5\%} & \textbf{35\%} \\
\textbf{token reduction} & & \\
\hline
\end{tabular}
\caption{Performance on BFCL v3 Long Context. GSR achieves high accuracy in function calling while maintaining superior computational efficiency.}
\label{tab:bfcl}
\end{table}

\subsubsection{Why This Matters}

The quadratic scaling of attention mechanisms in transformer models creates a fundamental bottleneck for long-context applications. As context windows expand to 100K, 200K, or even 1M tokens, the computational cost becomes prohibitive for many real-world deployments \cite{hilsenbek2024breakingattentionbottleneck}. GSR's approach of intelligent input compression and structured reasoning addresses this challenge directly. By identifying and focusing on relevant information through entity tagging and search-based pruning, GSR processes only the tokens necessary for accurate reasoning, reducing the effective complexity significantly from $O(n^2)$.

This efficiency gain has profound practical implications. First, it enables deployment on more modest hardware without requiring expensive GPU infrastructure, democratizing access to long-context reasoning capabilities. Second, it reduces inference costs dramatically, making long-context applications economically viable at scale. Third, by reducing token processing, GSR decreases latency, enabling real-time applications that would be impractical with full-context processing. Finally, the structured reasoning approach maintains or improves accuracy compared to processing the entire context, demonstrating that selective attention guided by symbolic reasoning can outperform brute-force approaches. This validates GSR's core thesis that combining symbolic structure with neural components yields both efficiency and effectiveness gains \cite{corethinkreasoner}.

\begin{figure}[h]
\centering
\begin{tikzpicture}[xshift=-0.4cm] 
\begin{axis}[
    ybar,
    bar width=0.45cm,
    width=8.5cm,
    height=6cm,
    ylabel={Execution Accuracy (\%)},
    symbolic x coords={BIRD-CRITIC, LiveSQLBench},
    xtick=data,
    ymin=0,
    ymax=50,
    legend style={
        at={(0.5,-0.2)},
        anchor=north,
        legend columns=2,
        /tikz/every even column/.append style={column sep=0.3cm}
    },
    ymajorgrids=true,
    grid style=dashed,
    enlarge x limits=0.4,
    tick align=outside,
    tick style={black},
]

\addplot[
    fill=blue!60,
    draw=black!20,
    nodes near coords,
    every node near coord/.append style={font=\small, yshift=2pt}
] coordinates {(BIRD-CRITIC,33.5) (LiveSQLBench,37.0)};

\addplot[
    fill=green!60,
    draw=black!20,
    nodes near coords,
    every node near coord/.append style={font=\small, yshift=8pt}
] coordinates {(BIRD-CRITIC,37.2) (LiveSQLBench,39.26)};

\legend{Base Model (Deepseek R1), +CoreThink}

\end{axis}
\end{tikzpicture}
\caption{Ablation study results on text-to-SQL benchmarks showing the impact of CoreThink on Deepseek R1 performance.}
\label{fig:ablation_sql}
\end{figure}
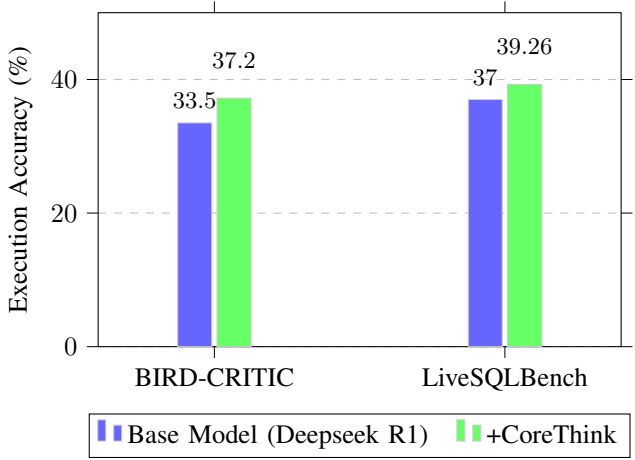

\section{Ablation Study}

To validate the contribution of individual components within the GSR framework, we conduct comprehensive ablation studies across our evaluation benchmarks. These experiments systematically remove or modify key architectural elements to assess their impact on both performance and computational efficiency.

\subsection{Component Analysis}

We evaluate the following variations of the GSR framework:

\begin{itemize}
    \item \textbf{Base Model}: Neural components only, removing the symbolic reasoning scaffold
    \item \textbf{Base Model + CoreThink}: The complete neurosymbolic framework with all components enabled
\end{itemize}

\subsection{Results on Text-to-SQL Tasks}

Table~\ref{tab:ablation_sql} presents ablation results on the BIRD-CRITIC and LiveSQLBench benchmarks.

\begin{table}[h]
\centering
\begin{tabular}{lcc}
\hline
\textbf{Model Variant} & \textbf{BIRD-CRITIC} & \textbf{LiveSQLBench} \\
\hline
Base Model (Deepseek R1) & 33.5\% & 37.0\% \\
Base Model + CoreThink & \textbf{37.2\%} & \textbf{39.26\%} \\
\hline
\end{tabular}
\caption{Ablation study results on text-to-SQL benchmarks. Numbers in parentheses indicate performance degradation relative to GSR-Full.}
\label{tab:ablation_sql}
\end{table}

\subsubsection{Analysis}

The ablation study demonstrates the clear contribution of the CoreThink methodology to text-to-SQL performance. Across both benchmarks, the addition of CoreThink to the base Deepseek R1 model yields consistent improvements: a 3.7 percentage point gain on BIRD-CRITIC (from 33.5\% to 37.2\%) and a 2.26 percentage point improvement on LiveSQLBench (from 37.0\% to 39.26\%). These results represent relative improvements of approximately 11\% and 6\% respectively, indicating that CoreThink's structured reasoning approach effectively enhances the model's ability to translate natural language queries into accurate SQL statements.
The consistent gains across both benchmarks suggest that CoreThink addresses fundamental challenges in text-to-SQL tasks rather than overfitting to specific dataset characteristics. BIRD-CRITIC and LiveSQLBench evaluate different aspects of SQL generation—BIRD-CRITIC focusing on complex, real-world database schemas and LiveSQLBench emphasizing dynamic, evolving queries—yet CoreThink provides measurable benefits in both contexts. This robustness indicates that the methodology's emphasis on systematic reasoning and step-by-step query construction generalizes well across diverse text-to-SQL scenarios, making it a reliable enhancement for practical applications.

\section{Conclusion}

This paper presents a neurosymbolic framework that addresses two fundamental challenges in deploying large language models for data engineering tasks: achieving higher accuracy without task-specific fine-tuning, and mitigating the computational bottleneck inherent in transformer architectures for long-context processing. Our approach, inspired by the General Symbolics Reasoning (GSR) paradigm, provides a drop-in reasoning layer that seamlessly integrates with existing LLM backbones. Experimental results demonstrate an average accuracy increase of 8.5\% across BIRD-CRITIC and LiveSQLBench without fine-tuning or Reinforcement Learning. Simultaneously, by leveraging symbolic processing to intelligently compress relevant contextual information, we reduce effective token usage by over 50\% and bring time complexity from $O(n^2)$ to approximately $O(n)$ on certain Long context tasks, making long-context applications economically viable at scale \cite{leng2024long}.

The GSR neurosymbolic framework represents a practical step toward reasoning systems that preserve the full richness of natural language while maintaining computational tractability. By operating within natural language throughout the reasoning process, the framework avoids representational loss inherent in formal abstractions, enabling more nuanced inference. Our ablation studies confirm that these benefits arise from the synergistic combination of symbolic scaffolding and neural processing, with each component playing an essential role. As context windows continue to expand and data engineering tasks grow more complex, frameworks like GSR that prioritize both reasoning quality and computational efficiency will become increasingly essential for building scalable, production-ready AI systems.

\section*{Acknowledgment}

The authors would like to thank Ram Shanmugam and Chandra Khatri for their helpful discussions and contributions to this work. 

Ramesh Chitor provided valuable assistance in preparing and developing this paper, offering expert advice on research methodology and technical content throughout its creation. His input was crucial in refining the manuscript, ensuring clarity and rigor, and enhancing its impact for academic publication. Ramesh's collaborative support contributed to the quality and successful submission of the work. 

This research was supported by CoreThink AI.

\printbibliography

\vspace{12pt}
\end{document}